\pdfoutput=1

\documentclass[11pt]{article}
\usepackage{multirow}
\usepackage[dvipsnames,table]{xcolor}
\usepackage{colortbl}
\usepackage{acl}
\usepackage{outlines}
\usepackage{times}
\usepackage{latexsym}
\usepackage{booktabs}
\usepackage{array}
\usepackage{tcolorbox}
\usepackage{geometry}
\usepackage{multicol}
\usepackage[T1]{fontenc}
\usepackage[utf8]{inputenc}

\usepackage{microtype}
\usepackage{todonotes}
\usepackage{inconsolata}
\usepackage{graphicx}

\usepackage{graphicx}
\usepackage{tikz}
\usepackage{listings}
\usepackage{xspace}
\usetikzlibrary{tikzmark}

\definecolor{darkgreen}{rgb}{0.0, 0.5, 0.0}
\newenvironment{chat} {
    
    \newcommand\who[1]{\par\hangindent=2.5em\hangafter=1 ##1:}
    \par\vskip2em
}{%
    \par
    \hangindent=0pt\hangafter=0
    \vskip2em
}

\definecolor{codegreen}{rgb}{0,0.6,0}
\definecolor{codegray}{rgb}{0.5,0.5,0.5}
\definecolor{codepurple}{rgb}{0.58,0,0.82}
\definecolor{backcolour}{rgb}{0.95,0.95,0.92}

\lstdefinelanguage{promptlanguage}{
    morecomment=[l][\color{codepurple}]{//},
    morecomment=[s][\color{blue}]{\{\{}{\}\}},
}
\lstdefinestyle{promptstyle}{
    backgroundcolor=\color{white},   
    commentstyle=\color{codegreen},
    keywordstyle=\color{magenta},
    numberstyle=\tiny\color{codegray},
    stringstyle=\color{codepurple},
    basicstyle=\ttfamily\small,
    frame = single,
    breakatwhitespace=false,         
    breaklines=true,                 
    captionpos=b,                    
    keepspaces=true,                 
    numbers=left,          
            xleftmargin=0.5cm,
        xrightmargin=0.5cm,
    numbersep=5pt,                  
    showspaces=false,                
    showstringspaces=false,
    showtabs=false,                  
    tabsize=2
}
\newcommand{\methodname}{RareScale\xspace}

\title{Rare Disease Differential Diagnosis with Large Language Models at Scale:\\ From Abdominal Actinomycosis to Wilson's Disease}

\author{%
  Elliot Schumacher\thanks{elliot@curai.com}
  \and
 Dhruv Naik
  \and
  Anitha Kannan\\
  Curai Health\\
}

\begin{document}
\maketitle
\begin{abstract}

Large language models (LLMs) have demonstrated impressive capabilities in disease diagnosis. However, their effectiveness in identifying rarer diseases, which are inherently more challenging to diagnose, remains an open question. Rare disease performance is critical with the increasing use of LLMs in healthcare settings.  This is especially true if a primary care physician needs to make a rarer prognosis from only a patient conversation so that they can take the appropriate next step. To that end, several clinical decision support systems are designed to support providers in rare disease identification. 
Yet their utility is limited due to their lack of knowledge of common disorders and difficulty of use.  

In this paper, we propose \methodname to combine the knowledge LLMs with expert systems.  We use jointly use an expert system and LLM to simulate rare disease chats.  This data is used to train a rare disease candidate predictor model.  Candidates from this smaller model are then used as additional inputs to black-box LLM to make the final differential diagnosis. Thus, \methodname allows for a balance between rare and common diagnoses.  We present results on over 575 rare diseases, beginning with Abdominal Actinomycosis and ending with Wilson's Disease.  Our approach significantly improves the baseline performance of black-box LLMs by over 17\% in Top-5 accuracy. We also find that our candidate generation performance is high (\textit{e.g.} 88.8\% on gpt-4o generated chats).
\end{abstract}

\section{Introduction}
\label{sec:intro}
\begin{figure*}[tb]
    \centering
    \includegraphics[width=0.85\linewidth]{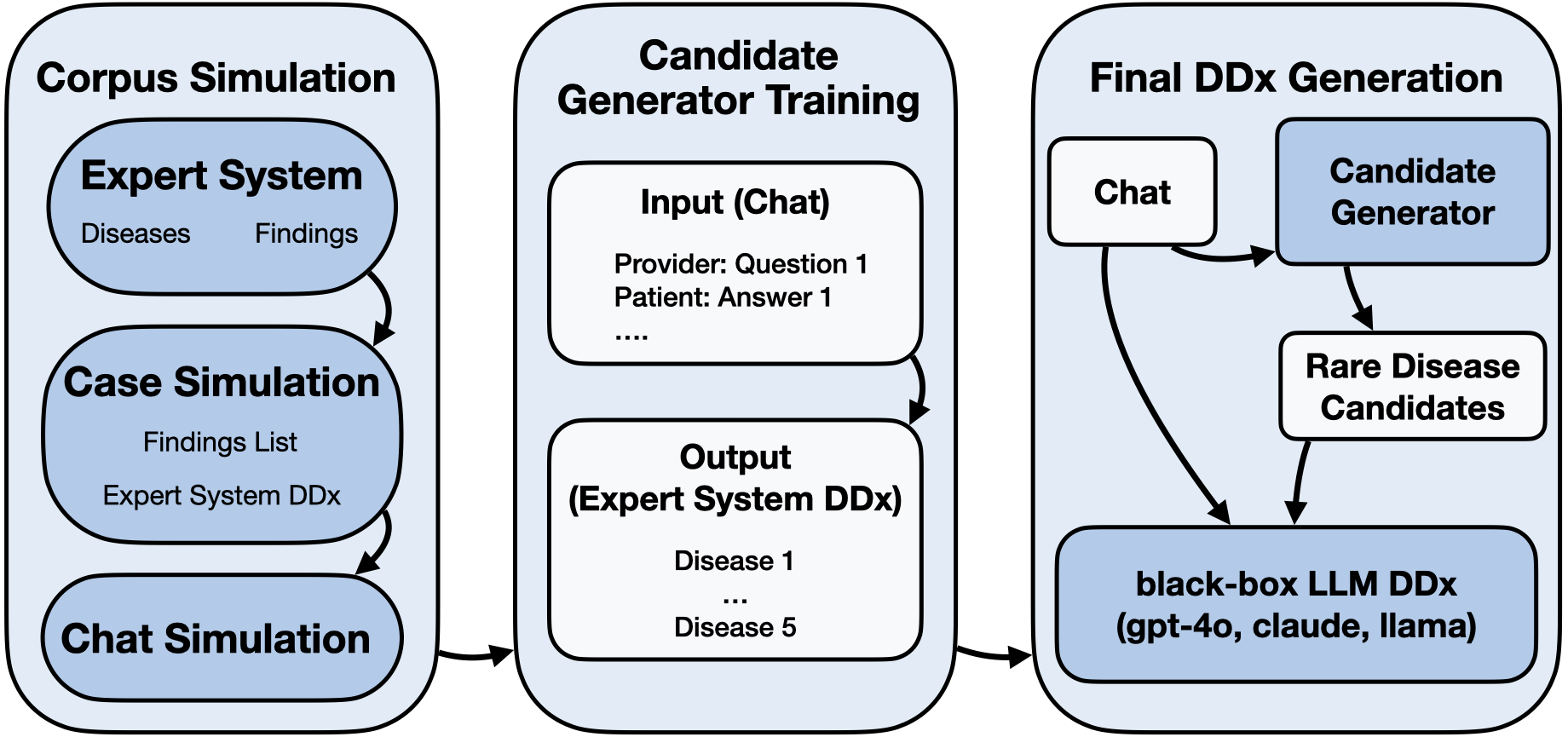}
    \caption{An overview of \methodname. Our approach consists of three stages.  First, we simulate a corpus of rare disease chats ( \S \ref{sec:data_gen} and Figure \ref{fig:expert_sim}).  Then, we train a candidate generation LLM (\S \ref{sec:candidate_gen_training}).  Finally, we perform inference to generate a final DDx (\S \ref{sec:ddx_gen}).}
    \label{fig:overview}
\end{figure*}

Rare diseases, often overlooked in the medical landscape, impact approximately 5.9\%  of the global population. For those affected, the journey to diagnosis is fraught with challenges, with an average wait of four to five years. Patients typically endure three misdiagnoses and consultations with at least five doctors during this process. This prolonged diagnostic odyssey is often due to limited provider knowledge, implicit biases, or symptom overlap with common conditions \cite{Kliegman17, Gao2021}. Only in a few instances does the absence of specific tests or the extreme rarity of the disease present the challenge.

Clinical decision support systems focusing on rare disease diagnoses emerged in the 1980s as a tool to address this challenge \citet{miller_internist-1_1982, mycin, barnett_dxplain:_1987, ExpertSystemsProbabilities, JacksonExpertSystems}. Practical use of these systems, however, has been constrained by several factors \cite{Miller94}, including lack of integration with physician workflows. Additionally, they are often custom-built explicitly for rare diseases and exclude common conditions.

Recent advances in large language models (LLMs; \citet{openai2024gpt4ocard, TheC3,grattafiori2024llama3herdmodels,jiang2023mistral}) achieve state-of-the-art performance on a variety of tasks \cite{zheng2023MT,wang2019super,hendrycksmeasuring}, including in high-stakes healthcare applications \cite{chen2023meditron70bscalingmedicalpretraining,thawakar-etal-2024-xraygpt,nair-etal-2024-dera}.  
Studies on rare disease diagnosis using LLMs show similar promise but have been on smaller, curated datasets of clinical vignettes that often include laboratory and imaging information to diagnose \cite{hu2023can,shyr2024identifying,sandmann2024systematic,chen2024rarebench,yang2024rdmaster,yang2024rdguru,mehnen2023chatgpt,shyr2024identifying,Olmo2024AssessingDD}.

Furthermore, the challenge with these studies is that 
the first point of contact for the patient is the primary care physician. For an untrained physician in rare diseases, the overlap of symptoms (\textit{e.g.}, \textit{pulmonary legionellosis} usually manifests with symptoms similar to flu or pneumonia)  makes it easy to overlook without specialized knowledge. While rare diagnoses often require additional information beyond the initial symptom presentation, they cannot be diagnosed if not considered in the first place.   This illustrates the criticality of including relevant rare diseases in the differential diagnosis, even before any lab work, directly based on patient-provider history-taking conversations. We found that applying black-box LLMs to the task of rare disease diagnosis shows only moderate performance (\textit{e.g.} 56.8\% Top-5 accuracy using gpt-4o).  This indicates that while LLMs have some knowledge of rare diseases, they also may struggle in certain cases.

In this paper, we improve the diagnostic capability of LLMs in identifying rare diseases based on two key insights. First, \textbf{given LLMs have some knowledge of rare diseases, can we more explicitly target this knowledge?}  Second, \textbf{can knowledge from rare expert systems be used to better evaluate and inform LLMs?}

We propose \methodname, which is designed to improve differential diagnoses of rare diseases directly from history-taking dialogues.  Figure \ref{fig:overview} provides an overview of our approach. 

We utilize a rare disease expert system and a medical case simulator to generate a broad set of structured clinical cases with closed-world assumptions within the expert system. These cases consist of structured findings, that a patient with the given rare disease may exhibit. These cases are then used as inputs to the history-taking conversation simulator that generates free-form provider questions and patient answers that conform to those cases. This approach covers 575 rare diseases, beginning with Abdominal Actinomycosis and ending with Wilson's Disease. 

Given that the expert system excludes common diseases, we develop a rare disease candidate generation model instead of a full diagnosis system. For each case, we generate a list of candidate rare diseases to prompt a larger, black-box LLM, which combines its general diagnostic capabilities with the specialized knowledge of the smaller model.

Our results show statistically significant performance improvements compared to relying solely on LLMs like GPT-4o.  For example, we find that the Top-5 accuracy improves from 56.7\% to 74.1\% on gpt-4o generated chats.  We believe this illustrates the potential of integrating curated expert knowledge sources and the power of LLMs.

\section{Problem Setup}
\label{sec:problem_statement}

When a patient visits a medical provider with a new, unknown health condition, the first step is understanding the issue deeply. This process, known as history-taking, consists of collecting all relevant information about the patient's current health issue, including positive and negative findings (or symptoms), via a series of questions. These details are used to make treatment decisions if a confident diagnosis can be made. Alternatively, they inform the selection of tests or imaging to help further diagnose. 

What if a possible diagnosis is not even considered because it's so rare, and the medical provider is unaware of it? 
This paper aims at proposing a method \methodname as a step towards removing that educational barrier. Given the history-taking conversation between the provider and the patient, the goal is to identify possible rare diseases that need to be considered.  When doing this, we must balance the fact that the disease is rare and that the patient is more likely to have a common disease.

\noindent {\bf Overview of the approach}  Figure \ref{fig:overview} provides the overview of \methodname: It trains a smaller expert LLM \S~\ref{sec:cand_gen} to generate candidate rare diseases using a simulated labeled conversational dataset using a combination of expert systems and LLMs (\S~\ref{sec:data_gen}). The candidates generated from this model are used as additional inputs to the black-box LLM to enable better weighing rare and common diseases. This leads to improved efficacy over using only black-box LLMs.

\begin{figure*}[tb]
    \centering
    \includegraphics[width=0.85\linewidth]{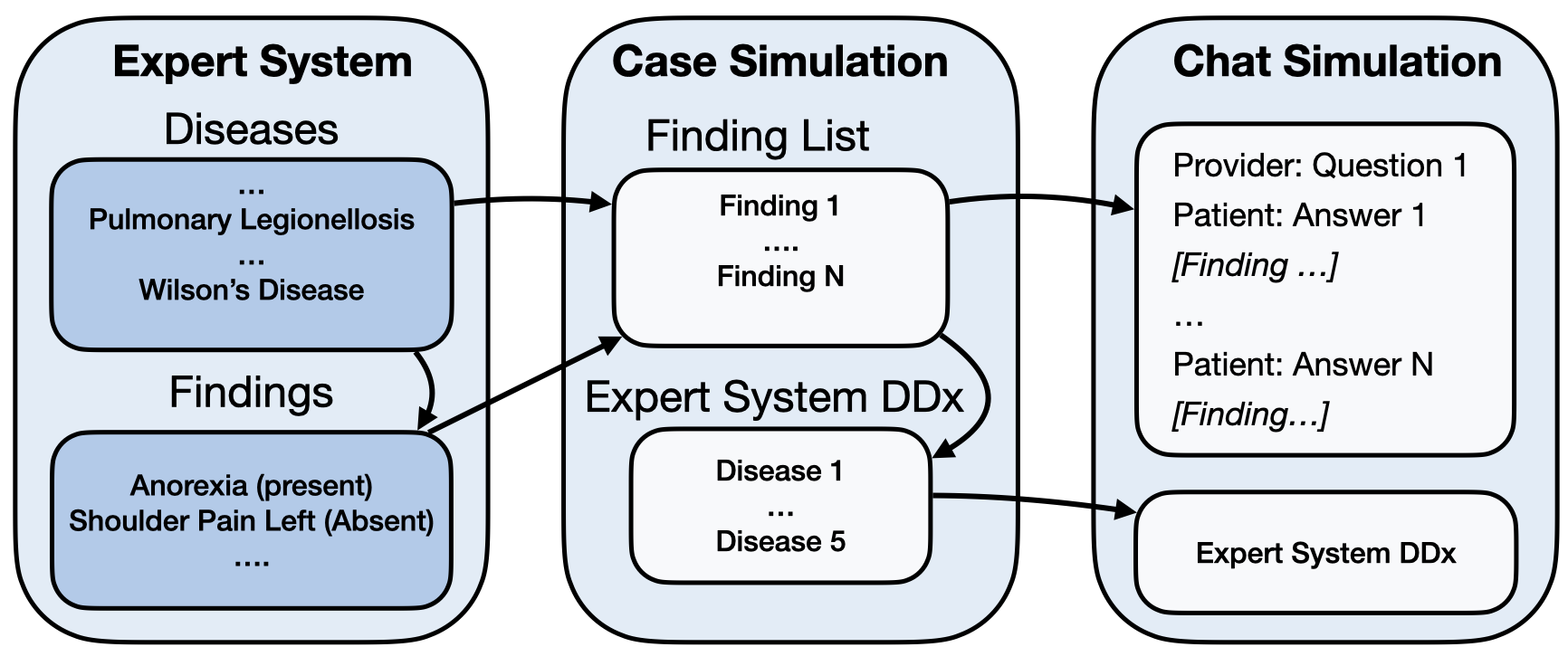}
    \caption{\methodname's Corpus Simulation Pipeline (\S \ref{sec:data_gen}).  Using an expert system, we create a set of structured case simulations, which are then used to guide an LLM in history taking chat generation.}
    \label{fig:expert_sim}
\end{figure*}

\section{Corpus Simulation}\label{sec:data_gen}

The use of synthetic data from large language models has been widely studied \cite{li-etal-2023-synthetic,yu2023training,liu2024best}. However, our task requires generating history-taking conversational data with labeled differential diagnosis. Since identifying differential diagnosis is the task we are trying to solve with \methodname, we took a different approach to generate labeled history taking chats to avoid simply evaluating an LLM based on it's existing knowledge.

As shown in Figure, \ref{fig:expert_sim}, we first simulate an example using expert system knowledge.  We start with a seed disease and generate a structured patient case (represented as a list of findings). 
Such a generation can have ambiguity about the final diagnosis. Therefore, we use the same expert system to assign a full differential diagnosis. The structured case is then used to simulate a history-taking conversation as detailed in \S ~\ref{sec:chat_generation} so that LLM needs to only generate a patient-facing question with the provided finding, while maintaining the conversational flow. We repeat this process independently for all diseases. The resulting dataset of history-taking conversation and differential diagnosis pairs is used to train rare diseases candidate generation model (\S~\ref{sec:cand_gen}).

\subsection{Generation of structured cases}
\label{sec:vignette}

We use a rare disease expert system, evolved from Internist-1 \cite{miller_internist-1_1982} and QMR \cite{miller_qmr}, to generate structured cases using expert-curated diagnostic rules based on a knowledge base of diseases, findings, and their relationships. Findings include symptoms, signs, lab results, demographics, or medical history. In this study, we focus on patient-answerable findings and {\it only} focus on symptoms. Each finding-disease link is defined by \emph{evoking strength} and \emph{frequency}, scored from 1 to 5 by a team of medical experts based on the studies for that disease. \emph{Evoking strength} measures the association between a finding and a disease, while \emph{frequency} indicates how often a finding occurs in patients with the disease. Each finding also has a disease-independent \emph{import} variable, indicating its global importance.

Our simulation algorithm, adapted from medical case simulator \citet{qmr-simulated-cases, pmlr-v85-ravuri18a}, starts by sampling a seed disease and sequentially constructing a set of findings. First, demographic variables are sampled, followed by predisposing factors, and then other findings are made to decrease frequency relative to the disease. Each finding is randomly determined to be present or absent, with impossible findings excluded and high co-occurrence findings prioritized. The simulation concludes once all findings in the knowledge base for that disease are considered, operating under a closed-world assumption that limits diseases and findings to those in the knowledge base.  Unlike previous approaches to simulation, we also use the expert system to compute differential diagnosis after the first six findings are sampled.  We use this to consider findings from other diagnoses in the differential diagnosis to prioritize sampling additional negative findings that overlap with the seed disease. This allows the sample to be slightly more targeted at the seed disease. In turn, we widen the gap between the seed disease and the remainder of the differential diagnosis.

Each simulation includes a differential diagnosis of up to 5 diseases as ranked by expert system scoring. A DDx may consist of multiple diseases in cases where a final diagnosis may not be ascertained without laboratory testing. We maintain this differential diagnosis list for training in \S \ref{sec:cand_gen}. We iterate through 630 rare diseases in the expert system, using each as a seed disease. We keep the generated structured case if the seed disease is top-scoring in the differential to reduce the effect of the noise in the simulation process. We set a minimum of $50$ valid simulations out of $200$ attempts. Any diseases falling below that are excluded (\textit{e.g.} Friedreich's ataxia), as this typically indicates that the disease is unidentifiable from symptoms alone. This results in 575 diseases in our dataset. We include an example in Appendix \S \ref{fig:chat_simulation_ex_app}.

\subsection{Generation of chats from structured cases}
\label{sec:chat_generation}
For each (structured case, DDx) pair, we start with creating a complete demographic profile.  This includes name, gender, age, race, education, and location by first selecting from the structured case.  If not present in the case, we randomly generate from the LLM. We incorporate location and education to introduce additional variability in patient language. This diversifies the profiles and reduces sensitivity to names and locations across different simulations, which could lead to misleading patterns.

We use the structured case along with the expanded demographic profile to anchor LLM-powered history-taking chat simulation. We prompt an LLM to create a conversation that closely adheres to that case.
The LLM-based chat simulator has access to the disease name, finding set, and demographic profile.  For each finding, we include a definition if the expert system provides one so the LLM's understanding of the term aligns with the expert system.  For each disease, we persist a database of messages that correspond to each finding (\textit{e.g.} ``I feel hot'' corresponds to \textit{fever (present)}). We include the previous message in the prompt for each subsequent generation and ask the LLM to generate a different one so it does not repeat.

We enforce a format for the chat, which begins with a "system" message with the patient's demographic information. The provider simulator starts the conversation open-ended, and the patient simulator reports the symptoms drawn from the findings set. For each patient utterance, the LLM outputs both the message and which findings are included.  This is to encourage the message to be consistent with the findings. We prompt LLM  for the patient simulator to report only at most, three findings 
in a single message and also prompt the LLM  for the provider simulator not to generate leading questions.

For the simulation, we use two different LLMs to generate our chats -- OpenAI's Gpt-4o and Anthropic's Claude-3.5-sonnet -- and use the same approach unless noted.  All components of the generation pipeline use only one of the LLMs. For chats generated by GPT-4o, we found that doing this process in a single prompt is sufficient (see Appendix Prompt \ref{prompt:chat_gpt4o}).  For Claude, we found that this generated chats which were very terse.  Therefore, we prompt Claude with the same information and ask it to generate one patient-provider turn at a time  (see Appendix Prompt \ref{prompt:chat_claude}).  The findings provided to Claude are separated into findings already included and those that need to be added.  

For chats generated by both LLMs, we finally run a checker prompt that ensures that the resulting chats include all symptoms in the finding set and edits the chat if not all are included.  If the first edit fails, we try two more times.  If all other attempts fail, we exclude this chat from our dataset.  This results in a corpus of rare disease chats -- 28,589 using GPT-4o and 14,573 using Claude (further statistics are included in Appendix Table \ref{tab:data_details}).  We include a full example chat using GPT-4o in Appendix Figure \ref{fig:example_gpt4o} and using Claude in Figure \ref{fig:example_claude}.

\section{Rare Disease Differential Diagnoses}\label{sec:cand_gen}
The \methodname scalable synthetic generation pipeline provides us with a sizable dataset of 575 rare disease history-taking conversations.  However, our expert system doesn't provide information on a variety of common diseases, such as the Common Cold.  A real-world diagnostic system will need to account for these diseases.  Given the high diagnostic performance of many LLMs \cite{generalistMedicalLangaugeModel, Rutledge2024DiagnosticAO,zhou2024largelanguagemodelsdisease,tu2024conversationaldiagnosticai}, it would be challenging to meet their performance even with a corpus of similar non-rare chats.

We therefore combine existing LLM diagnostic performance with a specialized rare-disease model.  To do this, we train a rare disease candidate generation system which proposes a set of at most 5 rare diseases.  This list is recall-oriented, to allow several possibilities to be considered.  These rare disease candidates are then passed through a prompt of a larger LLM to produce a final diagnostic list.  The combination can leverage the niche expertise of an expert system while also leveraging the broader knowledge of general LLMs.

\subsection{Candidate Generation} \label{sec:candidate_gen_training}
To generate rare disease candidates for a given history-taking conversation, we train a smaller LLM to generate a list of at most 5 rare diseases. We train using Llama 3.1 7b as our base model \cite{grattafiori2024llama3herdmodels} using the chat text as input (training details in  Appendix \S \ref{sec:training_details}).   We exclude the structured findings and all other information used in the previous section. Note that these history-taking chats do not include a final diagnosis from the provider, so the model must make inferences from the chat alone. As the target output, we use the differential diagnosis list produced by our expert systems which contains up to five diseases.  We only include the disease names ordered by likelihood per the expert system score but do not include the scores themselves.

\subsection{Diagnosis Generation}\label{sec:ddx_gen}

For the final step of generating the differential diagnosis for the conversation,  we use a larger black-box general LLMs\footnote{We explored using the publicly-available MEDITRON-70B\cite{chen2023meditron70bscalingmedicalpretraining}, but it could not follow the DDx instructions in the prompt, instead outputting unrelated text.} (gpt-4o, claude, llama 3.3 70b) to fuse common and rare diseases seamlessly. We leave the role of identifying the relevant common diseases to the LLM, but infuse the candidate rare diseases through inferencing on the smaller model we trained according to \S~\ref{sec:candidate_gen_training}

The prompt (Appendix Prompt \ref{prompt:ddx})  uses multi-stage instructions, instructing the LLM to generate a list of 5 diagnoses without considering the rare list.  Then, it is instructed to consider the rare disease candidate list. Finally, it selects whichever from the two sections is most appropriate, including discarding all rare candidates if best. The prompt is also flexible, so we can remove instructions for incorporating candidate rare diseases or include other means of candidate rare diseases when available. We study these variations in \S~\ref{sec:eval_ddx}.

\section{Evaluation and Results}

\begin{table*}[tb]
    \centering
    \begin{tabular}{l | c c c | c c c} \toprule
        \multirow{2}{*}{\textbf{ Differential Diagnosis}} & \multicolumn{3}{c|}{\textbf{gpt-4o test set (n=3403)}} & \multicolumn{3}{c}{\textbf{claude test set (n=2868)}} \\ \cmidrule(r){2-4} \cmidrule(l){5-7}
        & \textbf{Top-5} & \textbf{Top -1} & \textbf{MRR} & \textbf{Top-5} & \textbf{Top -1} & \textbf{MRR} \\ \midrule
        \textbf{baseline} & 56.80\% & 28.65\% & 0.390 & 56.69\% & 30.65\% & 0.406 \\ 
        \textbf{gpt-4o rare candidates} & 52.66\% & 25.95\% & 0.357 & 55.47\% & 29.04\% & 0.388 \\ 
        \textbf{\methodname candidates} & \textbf{74.38\%} & \textbf{33.12\%} & \textbf{0.471} & \textbf{71.41\%} & \textbf{33.23\%} & \textbf{0.461} \\ \bottomrule
    \end{tabular}
    \caption{Performance on generated gpt-4o ddx task. All metrics for \methodname on both datasets (see bolded) are significant using a two-sided Wilcoxon signed-rank test with $p<0.01$ compared to the no candidates baseline.}
    \label{tab:ddx}
\end{table*}

\begin{table*}[tb]
\centering
\begin{tabular}{l|cccccc} \toprule
\textbf{DDx LLM} & \textbf{Exact} & \textbf{Extremely Rel.} & \textbf{Relevant} & \textbf{Somewhat Rel.} & \textbf{Unrelated} \\ 
\midrule
\textbf{baseline gpt-4o} & 22.8\% & 19.9\% & 4.9\% & 21.0\% & 31.3\% \\ 
\textbf{\methodname gpt-4o} & 55.8\% & 8.8\% & 2.3\% & 12.8\% & 20.2\% \\ \midrule

\textbf{baseline claude} & 19.2\% & 16.9\% & 3.9\% & 14.5\% &45.6\%  \\ 
\textbf{\methodname claude} & 56.8\% & 10.7\% & 1.6\% & 10.6\% & 20.4\% \\ \midrule

\textbf{baseline Llama 3.3 70b} & 20.3\% & 19.3\% & 5.3\% & 21.7\% & 33.5\% \\ 
\textbf{\methodname Llama 3,3 70b} & 47.3\% & 12.2\% & 3.3\% & 15.4\% & 21.9\% \\ \bottomrule
\end{tabular}
\caption{We compare LLM baseline DDx generation performance to LLMs with addition of \methodname candidates.  We report the LLM as judge results across several categories of similarity, ranging from Exact Match to Unrelated. We combine gpt-4o and claude test sets for this analysis.}
\label{tab:ddx_by_sim}
\end{table*}

\begin{table*}[tb]
    \centering
    \begin{tabular}{l | c | c c c | c c c}
        \toprule
        \multirow{2}{*}{\textbf{Training Dataset}} & \multirow{2}{*}{\textbf{Training Size}} & \multicolumn{3}{c|}{\textbf{gpt-4o test set (n=3403)}} & \multicolumn{3}{c}{\textbf{claude test set (n=2868)}} \\ \cmidrule(r){3-5} \cmidrule(l){6-8}
         & & \textbf{Top-5} & \textbf{Top-1} & \textbf{MRR} & \textbf{Top-5} & \textbf{Top-1} & \textbf{MRR} \\ \midrule
        \textbf{claude} & 8837 & 48.37\% & 34.12\% & 0.4007 & 64.92\% & 45.64\% & 0.5371 \\ 
        \textbf{gpt-4o} & 21782 & 88.04\% & 63.88\% & 0.7410 & 44.18\% & 28.45\% & 0.3490 \\ 
        \textbf{gpt-4o downsampled} & 8813 & 70.88\% & 47.90\% & 0.5742 & 37.20\% & 23.25\% & 0.2884 \\ 
        \textbf{gpt-4o + claude} & 30619 & 88.80\% & 64.21\% & 0.7463 & 77.82\% & 56.35\% & 0.6526 \\ 
        \bottomrule
    \end{tabular}
    \caption{Evaluation on the candidate generation task, with MRR, Top-5 and Top-1 Accuracy.  We evaluate on models only trained on claude data, gpt-4o data, and both, and evaluate separately on claude and gpt-4o test sets. We include a model trained on a downsampled set of gpt-4o data that approximates the size of the claude training set.}

    \label{tab:cand_gen}
\end{table*}

To understand the performance of \methodname, we frame our evaluation around three questions.   First, does the \methodname end-to-end approach improve rare disease diagnoses over black-box LLMs (\S \ref{sec:eval_ddx})?  
Second, how does our expert candidate generation model perform at producing an accurate list of rare diseases for each case (\S \ref{sec:eval_cand})? Finally, how accurate are the simulated chats as judged by experts (\S \ref{sec:eval_exp})?  We include details on dataset creation in Appendix \S \ref{sec:eval_app}.

\subsection{DDx Improvements with \methodname}
\label{sec:eval_ddx}

We compare the end-to-end \methodname to two baselines.  First, in \textbf{base gpt-4o}, we generate a DDx with no external candidate list. \textbf{gpt-4o rare candidates} uses the \methodname prompt described in \S~\ref{sec:ddx_gen} without any candidate list. Similar to \methodname, \textbf{gpt-4o rare candidates} uses rare disease candidate list, but obtained using a separate LLM prompt. In contrast, \methodname uses a smaller trained model to generate the candidate list. For all three methods, we use the same prompt setup ( \S~\ref{sec:ddx_gen})  except for the additional candidate list and instructions. 

To be able to compare the generated DDx to the ground seed disease,  we use judge prompts adopted from \citet{tu2024conversationaldiagnosticai} (Appendix Prompts \ref{prompt:judge_binary} and \ref{prompt:judge_classification}), with gpt-4o as judge model. Previously, these prompts were shown to be well-calibrated with the human expert as a judge.  The first judge prompt returns a binary judgment, iterating through the DDx list and taking the first match present, if any.  The second judge prompt compares the full DDx list to the seed disease and returns a degree of similarity (unrelated to exact match) We use Top-1 accuracy, Top-5 accuracy, and mean reciprocal rank (MRR) as the metrics.

\paragraph{Results} Table \ref{tab:ddx} shows the performance of the combination of our best-performing candidate generation model (gpt-4o + claude, see \S \ref{sec:eval_cand}) with gpt-4o serving as a DDx generator. We find that \methodname produces the best performance across the board, including statistically significant gains over the no candidate baseline. Notably, we also find that performance is worse when using gpt-4o as the rare disease candidate generator, reinforcing the utility of a specialized model.

We believe the most crucial improvement is seen in the Top-5 performance, as adding the rare disease candidate in the DDx would allow a provider to consider that disease.  The degree of improvement -- 17.42\% -- illustrates this impact. While we also see improvements in MRR performance, we also observe that the rare disease candidate is not very likely to be in the first position.  

Including the rare disease candidate but not over-relying on the candidate list is likely the preferred behavior (see \S~\ref{sec:eval_cand}).  In many cases, a common disease should be considered first, but the inclusion of a rare possibility can help a provider to best identify next steps.  Using only our smaller rare disease candidate generation model would likely overproduce rare diagnoses, whereas providing a black box LLM with options allows it to weigh the entire picture.  Alternatively, the LLM may not be able to properly diagnose a specific rare condition because it has no knowledge of it to begin with (see the analysis for further discussion).

In Table \ref{tab:ddx_by_sim}, we compare LLM performance on the differential diagnosis task with and without \methodname broken down by similarity level. We combine the gpt-4o and claude chats for this analysis. Since this prompt uses varying levels of granularity and compares the entire DDx, the distribution is different than with the binary prompt, which only compares diseases one-to-one. 

In the case of gpt-4o, we see broad improvements when adding \methodname.  This includes a 33\% improvement in the number of Exact Matches.  Yet we see even larger performance improvements when applying \methodname to claude-sonnet's DDx capabilities.  We see a 25.2\% reduction in unrelated diagnoses and a 37\% increase in exact matches.  Surprisingly, we also see similar performance gains on Llama 3.3 70b, including a 27\% increase in exact matches and an 11.7\% reduction in unrelated matches. While there is a gap between the closed- and open-weight models, the margin is small.

\paragraph{Analysis}   We break down the results of the \methodname gpt-4o dataset results in Table \ref{tab:ddx} into performance by disease hierarchical category as taken from our expert system.  A disease can align with one or more categories. The full results are shown in Appendix Table \ref{tab:ddx_by_cat}.  The group with the largest number of diseases -- Infectious diseases -- performs slightly less than the average.  This could potentially be due to the overall broader number of diseases in this category, many of them closely related.  Other large groups, such as \textit{Neoplastic disease}, \textit{Impaired cardiovascular function}, and \textit{non-infectious inflammatory diseases} have roughly average performance.  On the lower end, as expected,  \textit{Congenital disorders due to abnormal fetal development} performs worst at 53.3\% Top-5 accuracy, whereas \textit{Disorders of smooth muscle contraction and/or relaxation} performs best at 94.4\%, although both are small categories.

\subsection{\methodname Candidate Generation}
\label{sec:eval_cand}
We measure the performance of \methodname candidate generation by comparing it to the seed disease. The candidate generator outputs a list  of at most $5$ candidates, and we compare to the seed disease using exact match. As before, we use  MRR and Top-1 and Top-5 accuracy.  Since GPT-4o and Claude required different approaches to generate the chat data, we considered different dataset variations to understand differences in the performance. 

\paragraph{Results} Table \ref{tab:cand_gen} reports the performance of \methodname on candidate generation.
Using the combined training set during training leads to the highest performance -- 88.8\% Top-5 on gpt-4o, and 77.82\% on claude on the test sets.   When only trained on one of the data sources,  performance is best when applied to data from the same source.   In our initial experiments, we found that including roughly $40$ to $50$ training examples per disease would achieve the same performance as more samples (\textit{e.g.} 100), but fewer would result in worse performance.  

While the gpt-4o test set performs similarly on gpt-4o trained models and combined training data, the claude chat test set sees a sizable performance bump when using all training data.   Since our claude dataset is smaller in size, we hypothesize that the additional signal from the gpt-4o training examples helps improve performance on the claude test set.  To illustrate this, we train a model on only gpt-4o data but randomly (per-seed disease basis) downsampled to roughly the same amount of data in claude's training set.  We find a similar pattern -- the training data reduction (40.5\% of the original size) results in a 17\% drop in Top-5 accuracy, and similar reductions for other metrics.  This suggests the lower claude performance is due to dataset size instead of worse corpus simulation performance.

While we only consider exact matches in Table \ref{tab:cand_gen}, we also run the similarity judge prompt (Prompt \ref{prompt:judge_classification}) on the gpt-4o + claude trained model, gpt-4o test set results. In only 5.2\% of cases, the model returns a DDx list unrelated to the seed diagnosis.  In the remaining cases, non-exact matches have some degree of similarity. We also include validation results in Appendix Table \ref{tab:validation_performance}

\subsection{Expert Evaluation of Chats}\label{sec:eval_exp}
We evaluate whether our synthetically generated chats are possible manifestations of the seed rare disease. For this, we use an external reviewing service that provides medical annotators and use third-year medical students for this task as they have been recently trained in similar tasks. They can also use additional resources (\textit{e.g.} medical references) during annotation. We also included hard negative examples during annotations to serve as distractors. We include details in Appendix \S \ref{sec:eval_app_eval}.

\paragraph{Results}
On our annotation set of 651 cases (with 73 annotation-specific negatives), we found the overall agreement rate to be 88.6\% with a Cohen's kappa of 0.53.  For the expected positive cases, \textit{i.e.} the disease was indicated for that patient case during simulation, the agreement rate was 90.48\%.  For the negative cases, the agreement rate was 73.97\%.  The high agreement rate of the positive examples used for training illustrates the performance of our synthetic generation pipeline. We include additional discussion in Appendix Section \ref{sec:eval_app_results}.

\section{Deployment Considerations}

We show that \methodname significantly improves the ability to identify potential rare diseases over baseline LLM performance.  We also know that most patients at a primary care clinic are likely to have a common disease.  \methodname aims at balancing common diseases with rare diseases so that the primary care physician does not oversubscribe to the possibility of a rare disease. Such rare diseases are hard to rule out without expensive testing and added mental toll.

When deploying in practice, there are several possible approaches. For instance, a system could only include rare diseases in cases where common diseases are ruled out first.  For example, \textit{ pulmonary legionellosis} presents similarly to the flu or pneumonia, and a provider would likely treat one of the common conditions first.  But when flu or pneumonia management fails, they can use \methodname to consider rarer conditions.  Alternatively, \methodname could be used to gather more information from the patient and exclude rarer conditions more quickly.
We believe that this is an open problem that may require physician training and clinical testing.

\section{Related Work}
Using LLMs for diagnosis is increasingly powerful \cite{generalistMedicalLangaugeModel, Rutledge2024DiagnosticAO,zhou2024largelanguagemodelsdisease,tu2024conversationaldiagnosticai}.  Previous rare disease investigations \cite{hu2023can,shyr2024identifying,sandmann2024systematic,chen2024rarebench,yang2024rdguru,mehnen2023chatgpt,shyr2024identifying,Olmo2024AssessingDD} focus on generating diagnoses from smaller sets of vignettes, which are concise summaries of a patient's health issue. Vignettes are commonly used in narrow settings.  Comparatively, using the history-taking chat replicates a primary care setting where confirmatory testing has not been performed. Previous chat-related work focused on a much smaller set \cite{yang2024rdmaster}. 

Prompting methods such as chain of thought \cite{wei2022cot} may improve rare disease performance to a degree but cannot enable the integration of external knowledge.  Other techniques such as retrieval augmented generation \cite{lewis2020rag} are unlikely to capture the complex nuances of symptoms as stated by patients. LLM performance on emerging data has also been studied \cite{mitchell2021fast,gu2024model}.  A related approach is \citet{yao2024scalable}, which uses a smaller expert model, but their approach seeks to update the original model.  \cite{gupta-etal-2024-model} illustrates that model editing can also lead to the model losing knowledge.  Other work \cite{zhao2024wildhallucinations} studies LLM performance in other rare data situations, and studies the related issue of  hallucinations\cite{fabbri-etal-2022-qafacteval,min-etal-2023-factscore}. 

\section{Conclusion}
We propose \methodname as a multi-component approach to improve rare disease diagnosis.  Based on curated rare disease expert systems, we use a medical case simulator to generate structured cases and differential diagnoses for each rare disease.  In turn, we prompt black-box LLMs to generate history-taking conversations of these cases.  This allows us to benchmark the performance of LLMs on rare disease history-taking conversations.  We show that   explicitly training an LLM for disease candidate generation, which can then provide suggestions to a black-box LLM, achieves statistically significant performance improvements compared to baselines.

While this paper focused on the rare disease diagnosis, we hope \methodname can transfer to other settings.  For example, one could leverage a location-specific or population-specific expert system instead of using a rare-specific expert system. This could especially be useful in settings where the real-world disease distribution differs significantly and at the tail of the data distribution, which black-box training data can not corroborate.

Although our approach scales the number of rare diseases to 575, many rare diseases \cite{rare_disease_mass_list} are left unaddressed.  An inherent limitation of our approach is that expert systems rely on the labor of highly-trained medical experts who review medical literature and curate knowledge.  This allows us to use the expert system as a surrogate source of knowledge instead of directly integrating with rare disease literature.  Yet expanding an expert system to all rare diseases with human experts alone is likely impossible. Future work should focus on using medical literature with techniques beyond retrieval augmented generation, which likely struggles to capture the subtlety and relative importance of the various studies. Alternatively, explorations that rely on electronic health record systems could remove the need for the expert system while still using a smaller model tuned for the rare disease candidate generation to be used as in \methodname.

\section{Limitations}

One limitation of our work is that we are limited to the rare diseases and symptoms provided in our rare disease expert system.  While our approach does cover a broader set of rare diseases than existing work, it does not cover the vast set of diseases present in the world. For example, \cite{rare_disease_mass_list} lists over 10,000.  Yet rare diseases highly vary in prevalence, with only about 100 rare diseases accounting for 80\% of diagnoses\cite{evansRareGeneral}.  Others are so rare that only a few cases have ever been identified. Expanding to these extremely rare diseases is a remaining challenge. 

We are also limited by the number of expert annotations that are feasible for this task.  While the results were positive, we couldn't annotate a larger amount without spending excessively more time and money. Additionally, we are unable to release the datasets due to legal requirements.  While this is a common issue in the medical domain, we hope that future work can provide open datasets for broader use.

\bibliography{custom}
\clearpage

\appendix
\section{Appendix}
\label{sec:appendix}

\subsection{Training Details}\label{sec:training_details}
We train for 10 epochs using supervised fine-tuning only on 8 Nvidia A100s.  Each training run, using the huggingface transformers library (v4.43.3), took 2-3 hours depending on the setup. We used the AdamW 8-bit optimizer \cite{Loshchilov2017DecoupledWD}, a learning rate of $2e^{-05}$, and a batch size of 4. We selected these parameters based on early experiments on subsets of our larger disease set.

\subsection{Evaluation Details}
\label{sec:eval_app}
Following the approach in Section \ref{sec:data_gen}, we generate chats using gpt-4o and claude-sonnet 3.5.  For each, we create train, validation, and test splits as detailed in Appendix Table \ref{tab:data_details}.  We also provide the average and standard deviation of the number of findings and the number of messages.  To ensure that both the validation and test sets are distinct from the training set, we removed all examples from the training set that have the exact same structured case in the validation or test sets. We generated a smaller dataset for claude due to the additional cost and time of running multiple prompts per chat.

\subsubsection{Expert Evaluation}
\label{sec:eval_app_eval}

We provide the annotators with the chat, list of findings, and the seed disease.  They are asked
\textit{``Assume the patient has a rare disease and more common conditions have been ruled out. Is this specific rare disease a possible diagnosis for the patient? This rare disease doesn’t need to be the most likely diagnosis, but just a possible diagnosis.''}.  They are prompted to respond yes or no, and provide a 1-2 sentence explanation.  They are allowed to use any external reference material they choose.

One challenge in this annotation task is that our dataset does not include negative examples as they are not required for training.  However, only annotating positive examples may lead to annotators blindly answering ``yes''.  Therefore, we generate negative training examples solely for expert evaluation and ask annotators to review datasets that are 90\% positives and 10\% negatives.  

Generating a chat where a rare disease is not possible is a challenging task.  Providing annotators with cases that are obviously negative would not be informative. Generating cases where a disease is negative but close requires conclusively ruling out a disease, which is challenging in a history taking setting where lab work isn't performed. 
To achieve a close approximation, we regenerate examples from our expert system and take diagnoses that were present in earlier simulation phases but removed later when additional findings were added.  These ``discarded diagnoses'' are likely to be negative diagnoses for the findings. 

We separately generate chats for these findings and discarded diagnosis using the process described in Section \ref{sec:data_gen}.  As a second filtering step, we provide the chat and discarded diagnosis to a gpt-4o prompt, and prompt it to provide the same binary judgment and explanation as we do with the annotators.  Finally, we manually filter out chats where the explanation rests mostly on likelihood (\textit{i.e.} disease A is unlikely because disease B is more likely), and retain cases where there is a clear separation.  However, we want to emphasize that even in these cases, it is hard to completely rule out a rare disease.  All cases were reviewed by at least one annotator -- cases where the original annotator disagreed with the initial label was reviewed by a second labeler.

\subsection{Expert Evaluation Results}\label{sec:eval_app_results}

While the agreement rate for negative cases is lower than the agreement for positive cases, it is inherently challenging to create believably negative cases.  It also does not impact \methodname's performance given that we only add negatives to ensure that annotators do not blindly choose yes. We include a selection of annotator reasoning in Appendix Table \ref{tab:ann_reasoning}. In many cases where the annotator disagrees with our baseline label, they note that the full symptom set isn't present for the disease.  While that may make it unlikely, that does not mean it can be ruled out entirely because an atypical presentation could occur or further symptoms may arise.  This tension is inherent when dealing with rare diseases.

\begin{figure*}
\begin{tcolorbox}[boxsep=2pt,left=2pt,right=2pt,top=2pt,bottom=2pt,colback=White,]
\begin{multicols}{2}

\textbf{Findings List:}

Abdomen pain left upper quadrant (absent)\\
Abdomen pain non-colicky (absent)\\
Anorexia (present)\\
Chest pain lateral sharp (present)\\
Chest pain substernal exacerbation with cough (present)\\
Cigarette smoking history (present)\\
Dyspnea abrupt onset (present)\\
Headache severe (present)\\
Myalgia (present)\\
Onset abrupt (absent)\\
Rigors (present)\\
Shoulder pain left (absent)\\
\textit{larger }


\textbf{DDx List:}\\
pulmonary legionellosis \textbf{(seed disease)}\\
pneumococcal pneumonia\\
mycoplasma pneumonia\\
brucellosis

\columnbreak
\textbf{Patient-Doctor Chat:}\\
\textbf{System:} Patient is Tyler, a 24-year-old male.\\
\textbf{Doctor:} Hi Tyler, I'm here to help you today. Can you tell me what's been going on?\\
\textbf{Patient:} Hey doc, I've been feeling really off lately.\\
\textbf{Doctor:} Can you describe what you mean by off? Are you experiencing any pain or discomfort?\\
\textbf{Patient:} Yeah, there's this sharp pain on the side of my chest, and it gets worse when I breathe.\\
\textbf{[chest pain lateral sharp-present]}\\
\textbf{Doctor:} That sounds uncomfortable. Have you noticed any other types of chest pain, maybe when you cough?\\
\textbf{Patient:} Not really sharp pain, but it does hurt under my sternum when I cough.\\
\textbf{[chest pain substernal exacerbation with cough-present]}

\end{multicols}
\end{tcolorbox}
\caption{Example expert system simulation to chat}
\label{fig:chat_simulation_ex_app}
\end{figure*}

\begin{table*}
\centering
\begin{tabular}{p{0.5\textwidth}p{0.1\textwidth}p{0.1\textwidth}p{0.08\textwidth}p{0.08\textwidth}}
        \toprule
        \textbf{Dx Category} & \textbf{Top K Accuracy} & \textbf{Top 1 Accuracy} & \textbf{MRR} & \textbf{N} \\
        \midrule
        Congenital disorders due to abnormal fetal development & \cellcolor{blue!0}0.533 & \cellcolor{blue!34}0.433 & \cellcolor{blue!25}0.458 & \cellcolor{blue!0}30 \\
        Immune system disorders & \cellcolor{blue!7}0.548 & \cellcolor{blue!2}0.190 & \cellcolor{blue!10}0.295 & \cellcolor{blue!10}42 \\
        End organ damage secondary to other disorders & \cellcolor{blue!14}0.583 & \cellcolor{blue!0}0.083 & \cellcolor{blue!0}0.222 & \cellcolor{blue!7}36 \\
        Disorders with excess or abnormal fluid accumulation & \cellcolor{blue!22}0.603 & \cellcolor{blue!10}0.256 & \cellcolor{blue!12}0.370 & \cellcolor{blue!28}78 \\
        Disorders of the hematopoietic and lymphatic systems & \cellcolor{blue!34}0.667 & \cellcolor{blue!0}0.000 & \cellcolor{blue!10}0.306 & \cellcolor{blue!0}6 \\
        Drug induced injury alias adverse drug effects & \cellcolor{blue!40}0.699 & \cellcolor{blue!2}0.192 & \cellcolor{blue!15}0.361 & \cellcolor{blue!26}73 \\
        Degenerative disorders & \cellcolor{blue!45}0.704 & \cellcolor{blue!8}0.241 & \cellcolor{blue!20}0.391 & \cellcolor{blue!38}108 \\
        Infectious disease alias infections & \cellcolor{blue!49}0.711 & \cellcolor{blue!10}0.278 & \cellcolor{blue!25}0.424 & \cellcolor{blue!50}862 \\
        Neoplastic disease & \cellcolor{blue!50}0.721 & \cellcolor{blue!8}0.236 & \cellcolor{blue!22}0.384 & \cellcolor{blue!19}330 \\
        Disorders involving cysts stones or calculi & \cellcolor{blue!50}0.722 & \cellcolor{blue!43}0.611 & \cellcolor{blue!32}0.650 & \cellcolor{blue!0}18 \\
        Impaired cardiovascular function & \cellcolor{blue!50}0.738 & \cellcolor{blue!24}0.359 & \cellcolor{blue!32}0.486 & \cellcolor{blue!22}390 \\
        Disorders due to toxic or chemical or radiation injury & \cellcolor{blue!50}0.745 & \cellcolor{blue!31}0.398 & \cellcolor{blue!36}0.523 & \cellcolor{blue!27}98 \\
        Musculoskeletal disorders & \cellcolor{blue!50}0.750 & \cellcolor{blue!12}0.292 & \cellcolor{blue!30}0.467 & \cellcolor{blue!0}24 \\
        Non-infectious inflammatory disease & \cellcolor{blue!50}0.753 & \cellcolor{blue!21}0.326 & \cellcolor{blue!32}0.477 & \cellcolor{blue!22}384 \\
        Metabolic disorders & \cellcolor{blue!50}0.778 & \cellcolor{blue!21}0.333 & \cellcolor{blue!32}0.482 & \cellcolor{blue!38}144 \\
        Bleeding disorders and coagulopathies & \cellcolor{blue!50}0.783 & \cellcolor{blue!14}0.267 & \cellcolor{blue!28}0.449 & \cellcolor{blue!12}60 \\
        Disorders of thorax cardiovascular system and lymphatic ducts & \cellcolor{blue!50}0.783 & \cellcolor{blue!34}0.522 & \cellcolor{blue!30}0.590 & \cellcolor{blue!0}23 \\
        Endocrine disease & \cellcolor{blue!50}0.788 & \cellcolor{blue!30}0.417 & \cellcolor{blue!34}0.547 & \cellcolor{blue!35}132 \\
        Fibrosis or scarring of visceral organ & \cellcolor{blue!50}0.796 & \cellcolor{blue!8}0.245 & \cellcolor{blue!27}0.436 & \cellcolor{blue!8}49 \\
        Neuropsychiatric disorders & \cellcolor{blue!50}0.803 & \cellcolor{blue!35}0.439 & \cellcolor{blue!38}0.573 & \cellcolor{blue!10}66 \\
        Disorders secondary to trauma & \cellcolor{blue!50}0.806 & \cellcolor{blue!24}0.361 & \cellcolor{blue!35}0.519 & \cellcolor{blue!7}36 \\
        Impaired fluid flow within hollow viscus or viscera non-vascular & \cellcolor{blue!50}0.833 & \cellcolor{blue!43}0.583 & \cellcolor{blue!32}0.649 & \cellcolor{blue!0}24 \\
        Multisystem disorders & \cellcolor{blue!50}0.833 & \cellcolor{blue!50}0.667 & \cellcolor{blue!42}0.708 & \cellcolor{blue!0}6 \\
        Disorders associated with pregnancy & \cellcolor{blue!50}0.833 & \cellcolor{blue!34}0.542 & \cellcolor{blue!42}0.663 & \cellcolor{blue!0}24 \\
        Inherited congenital or degenerative disorders & \cellcolor{blue!50}0.847 & \cellcolor{blue!35}0.458 & \cellcolor{blue!30}0.588 & \cellcolor{blue!38}144 \\
        Miscellaneous mechanical disorders & \cellcolor{blue!50}0.861 & \cellcolor{blue!34}0.542 & \cellcolor{blue!36}0.655 & \cellcolor{blue!24}72 \\
        Kidney and urinary tract disorders & \cellcolor{blue!50}0.875 & \cellcolor{blue!35}0.458 & \cellcolor{blue!36}0.654 & \cellcolor{blue!0}24 \\
        Disorders due to mechanical tear or trauma or visceral erosion & \cellcolor{blue!50}0.875 & \cellcolor{blue!50}0.667 & \cellcolor{blue!50}0.764 & \cellcolor{blue!0}24 \\
        Disorders of abdomen digestive system and/or nutrition & \cellcolor{blue!50}0.889 & \cellcolor{blue!24}0.389 & \cellcolor{blue!38}0.573 & \cellcolor{blue!0}18 \\
        Disorders due to nutritional and/or vitamin deficiency & \cellcolor{blue!50}0.917 & \cellcolor{blue!43}0.611 & \cellcolor{blue!50}0.738 & \cellcolor{blue!7}36 \\
        Electrophysiological neurological disorders & \cellcolor{blue!50}0.917 & \cellcolor{blue!34}0.542 & \cellcolor{blue!42}0.687 & \cellcolor{blue!0}24 \\
        Disorders of smooth muscle contraction and/or relaxation & \cellcolor{blue!50}0.944 & \cellcolor{blue!50}0.722 & \cellcolor{blue!50}0.801 & \cellcolor{blue!0}18 \\
        \bottomrule
\end{tabular}
\caption{Performance from \methodname, gpt-4o in Table \ref{tab:ddx} broken down by disease category. Note that a disease may fall into multiple categories.}
\label{tab:ddx_by_cat}
\end{table*}

\begin{table*}
    \centering
    \centering
    \begin{tabular}{c | c c c c}
        \toprule
        \textbf{generation model} & \textbf{split} & \textbf{size} & \textbf{findings} & \textbf{messages} \\ 
        \midrule
        \multirow{3}{*}{\textbf{gpt-4o}} 
        & train & 21782 & 11.88 $\pm$ 1.91 & 17.54 $\pm$ 3.58 \\ 
        & val & 3404 & 11.69 $\pm$ 2.08 & 17.31 $\pm$ 3.78 \\ 
        & test & 3403 & 11.66 $\pm$ 2.05 & 17.27 $\pm$ 3.69 \\ 
        \midrule
        \multirow{3}{*}{\textbf{claude}} 
        & train & 8837 & 11.88 $\pm$ 1.90 & 14.54 $\pm$ 3.47 \\ 
        & val & 2868 & 11.69 $\pm$ 2.04 & 14.11 $\pm$ 3.38 \\ 
        & test & 2868 & 11.70 $\pm$ 2.08 & 14.30 $\pm$ 3.48 \\ 
        \bottomrule
    \end{tabular}
    \caption{Data statistics for train, validation, and test sets.  We include the number of chats, the average number of findings and standard deviation, and the average number of messages and standard deviation.}
    \label{tab:data_details}

\end{table*}

\begin{table*}
\small
\centering
\begin{tabular}{p{0.1\textwidth} p{0.1\textwidth} p{0.15\textwidth} p{0.47\textwidth}}
    \toprule
   Synthetic & Annotator & Disease & Reasoning \\ 
    \midrule
    yes & yes & histoplasma meningitis & This disease is possible because the patient’s history of lymphoma and transplant makes them immunocompromised, which increases their risk of histoplasmosis meningitis. The histoplasmosis meningitis could cause this patient's symptoms of nausea/vomiting, fever/chills, and severe headache. \\      \midrule
    yes & yes & cerebral malaria & This disease is possible because urinary and bowel incontinence, intractable headache, visual loss/retinopathy, and fever point to an infectious cerebral process that could be caused by cerebral malaria. \\    \midrule
    yes & yes & neurogenic osteoarthropathy alias charcot\_joint\_disease & This could be possible charcot joint disease, which is set off by trauma to a neuropathic extremity and can cause joint pain. Although Charcot's is technically most common in the foot, it can also extend to any major joint like the knees, shoulder, hip, etc. \\     \midrule
    yes & yes & glaucoma acute angle closure & Blurriness with rainbow rings/halos, nausea/vomiting, headache, decreased vision, and sudden onset are all symptoms of acute angle-closure glaucoma. \\    \midrule
    yes & yes & cytomegalovirus infection disseminated & This disease is possible because the patient is immunocompromised by their organ transplant, and their symptoms of vision changes, diarrhea, fever, chills, vomiting, and myalgias are consistent with a disseminated cytomegalovirus infection. \\     \midrule
    yes & no & cutaneous anthrax & Lack of bump/ulcer/eschar. though presence of pruritis and exposure to possible animals infected as a vet, symptoms are nonspecific. \\     \midrule
    yes & no & herpes zoster & This disease is not possible because although herpes zoster can have peripheral symptoms including this patient's myalgia, headache, and abdominal pain, this patient does not have the characteristic dermatomal rash, skin changes, or skin-level pain of herpes zoster. \\    \midrule
    no & no & pulmonary aspergillosis invasive type & This disease is not possible because the patient is not immunocompromised, has not had recent surgeries or pneumonia, or chemotherapy, and does not have a cough, which I would expect from pulmonary aspergillosis since it is an opportunistic pulmonary infection. \\     \midrule
    no & no & henoch schonlein syndrome alias henoch-schonlein purpura & This disease is not possible because the patient does not report the purpura around the legs/gluteus that is characteristic for Henoch Schonlein purpura. The patient is also not in the typical age group for this disease, which primarily is in pediatric populations. \\     \midrule
    no & yes & bronchial asthma & This disease is possible because of the patient's history of dyspnea at rest and worse with activity. The urinary frequency could also be related because there are positive associations between bronchial asthma and increased urge to urinate. \\     
    \bottomrule
\end{tabular}
\caption{Sample explanations from the annotation task. Note that the disease column indicates which disease they were asked to annotate against -- this is the expected disease for the positive synthetic labels, but is not expected for the negative ones.}
\label{tab:ann_reasoning}
\end{table*}

\begin{table*}
    \centering
    \begin{tabular}{c | c c c | c c c}
        \toprule
        & \multicolumn{3}{c|}{\textbf{gpt-4o chats}} & \multicolumn{3}{c}{\textbf{claude chats}} \\
        \cmidrule(lr){2-4} \cmidrule(lr){5-7}
        & \textbf{Top K} & \textbf{Top-1} & \textbf{MRR} & \textbf{Top K} & \textbf{Top-1} & \textbf{MRR} \\ 
        \midrule
        gpt-4o only & 88.28\% & 63.13\% & 0.7384 & 47.25\% & 31.80\% & 0.3817 \\ 
        claude only & 48.38\% & 34.43\% & 0.4025 & 65.93\% & 45.71\% & 0.5415 \\ 
        combined & 88.57\% & 64.10\% & 0.7444 & 79.64\% & 57.46\% & 0.6675 \\ 
        \bottomrule
    \end{tabular}
    \caption{Candidate generation metrics for validation set.}
    \label{tab:validation_performance}
\end{table*}

\begin{figure*}[p]
\begin{tcolorbox}[boxsep=2pt,left=2pt,right=2pt,top=2pt,bottom=2pt,colback=White,]
\textbf{gpt-4o Generated Chat for disseminated intravascular coagulation}
\vspace{-2em}
\begin{chat}
\who{System} Patient is George, a 67-year-old male. \\
\textit{[age greater than 55 (present) , sex male (present)]}
\who{Doctor} Hi George, I'm here to help you today. Can you tell me what's been going on?
\who{Patient} Well, it all started pretty suddenly. I've been feeling really off. \\
\textit{[onset abrupt (present)]}
\who{Doctor} I see. Can you describe some of the symptoms you've been experiencing?
\who{Patient} I've had a fever on and off. It's been bothering me. \\
\textit{[fever (present)]}
\who{Doctor} Alright, fever can be quite uncomfortable. Have you noticed any changes in your appetite?
\who{Patient} I haven't really felt like eating much lately. Just no appetite. \\
\textit{[anorexia (present)]}
\who{Doctor} Got it. Have you experienced any issues with bleeding or bruising?
\who{Patient} Yeah, even a small bump seems to make me bleed a lot. It's strange. \\
\textit{[bleeding excessive after minor trauma (present)]}
\who{Doctor} That does sound unusual. Have you had any nosebleeds recently?
\who{Patient} No, I haven't had any nosebleeds. \\
\textit{[epistaxis or epistaxis severe history (absent)]}
\who{Doctor} Okay, and how about your breathing? Any issues there?
\who{Patient} Yes, I've been feeling short of breath even when I'm just sitting around. \\
\textit{[dyspnea at rest (present)]}
\who{Doctor} That's important to note. Have you had any recent surgeries?
\who{Patient} Yes, I had surgery a few weeks ago. \\
\textit{[surgery recent history (present)]}
\who{Doctor} Thanks for sharing that. Have you experienced any vomiting?
\who{Patient} Yes, I've been throwing up, and sometimes it looks really dark, like coffee grounds. \\
\textit{[vomiting coffee ground (present) , vomiting recent (present)]}
\who{Doctor} I see. Have you ever had a reaction to a blood transfusion?
\who{Patient} Yeah, I've had a reaction before. \\
\textit{[transfusion reaction history (present)]}
\who{Doctor} Alright. Have you experienced any seizures?
\who{Patient} Yes, I've had some seizures. \\
\textit{[seizures grand mal single or intermittent (present)]}
\who{Doctor} And how about any back pain?
\who{Patient} No, my back's been fine. \\
\textit{[back pain midline severe (absent)]}
\end{chat}
\vspace{-2em}
\end{tcolorbox}

\caption{Example gpt-4o generated chat.  The simulated findings are in italics. Note that we do not provide the findings as input to train candidate generation model }\label{fig:example_gpt4o}

\end{figure*}

\begin{figure*}[p]
\begin{tcolorbox}[boxsep=2pt,left=2pt,right=2pt,top=2pt,bottom=2pt,colback=White,]
\textbf{Claude Generated Chat for porphyria acute intermittent}
\vspace{-2em}
\begin{chat}
\who{System} Sarah is a 35-year-old female in Seattle, Washington.
\\\textit{[sex female (present) , age 26 to 55 (present)]}
\who{Doctor} Hello Sarah, I'm Dr. Smith. What brings you to see me today?
\who{Patient} Hey doctor. I've been dealing with these terrible stomach pains lately. They come and go in waves, and they're really intense, especially in my upper right side. It started about two days ago.
\\\textit{[abdomen pain present (present) , abdomen pain acute (present) , abdomen pain colicky (present)]}
\who{Doctor} I'd like to know more about these pains. Have you had any pain in other parts of your stomach, like around your belly button or on the left side?
\who{Patient} No, not really. The pain is mostly just in my upper right side. It doesn't really go to my belly button area or the left side at all.
\\\textit{[abdomen pain periumbilical (absent) , abdomen pain left upper quadrant (absent)]}
\who{Doctor} How long have you been experiencing these stomach pains? Have you had similar episodes in the past months?
\who{Patient} No, this is really the first time it's happened. Like I said, it just started a couple days ago. I haven't had any ongoing stomach issues before this.
\\\textit{[abdomen pain chronic (absent)]}
\who{Doctor} Have you noticed any issues with eating? Any trouble swallowing certain foods?
\who{Patient} Actually, yeah. I've been having a hard time swallowing, especially with meat and bread. They seem to get stuck, and I have to really work to get them down. It's making eating pretty difficult.
\\\textit{[dysphagia solids (present)]}
\who{Doctor} Have you noticed any vision changes recently? Anything unusual when you're looking at objects?
\who{Patient} Yeah, actually it's weird - I keep seeing double. Like when I look at my computer screen at work, I see two screens. It's really making it hard to code.
\\\textit{[eyes diplopia (present)]}
\who{Doctor} Have you experienced any muscle cramps recently, like muscles tightening up and staying tight for a minute or longer?
\who{Patient} No, I haven't had any muscle cramps like that. The stomach pain is bad enough, but my muscles feel normal.
\\\textit{[muscle cramps (absent)]}
\who{Doctor} Could you tell me about your family background? Where are your ancestors from?
\who{Patient} Oh yeah, my family is from Sweden and Norway. Both my parents' families came from there a few generations back.
\\\textit{[ethnic background northern european (present)]}
\who{Doctor} Have you noticed any connection between your symptoms and your menstrual cycle?
\who{Patient} Now that you mention it, yes. I started my period right when these symptoms began. It seems like the pain and other symptoms got worse when my period started.
\\\textit{[menses precipitation or exacerbation of disease history (present)]}
\end{chat}
\vspace{-2em}
\end{tcolorbox}

\caption{Example claude generated chat.  The simulated findings are in italics. }\label{fig:example_claude}

\end{figure*}

\begin{lstlisting}[language=promptlanguage,float=*,caption=Prompt for generating chats with gpt-4o. ,label={prompt:chat_gpt4o}]
Create a history taking conversation where a patient has the following medical profile, and is chatting with a doctor.
Include only the history taking part of the chat, do not include any diagnosis or include the doctor confirming the findings.
Be sure to use language appropriate to the patient's profile, and do not make the patient text technical unless indicated.  Use informal terms when appropriate
Ensure all findings are explicitly discussed, both present and absent statuses.
At the end of the chat, all findings must be included in a previous message.

The first message should consist of a generic system message which  explicitly states the patient's name, age, gender (stated directly), race if present.  Do not include any findings beyond that list.
The second message should consist of a generic greeting from the doctor.  The doctor does not have any information at this point.
Do not include more than two findings in a question answer pair.
The patient is also unlikely to be able to explain their condition in a coherent manner.  Do not assume that they will clearly offer up the findings, and the doctor may have to ask multiple questions before they get to a finding.
The doctor should not ask leading questions.

Disease: {{disease}}

--Patient Profile--
{{demographics_str}}
--End Profile--

--Findings--
{{findings_str}}
--End Findings--

Output format in yaml
--Format--
chat:
 - role : "System"
   msg : [MSG ABOUT GENDER, AGE, RACE]
   findings :
    - [FINDING 0 , STATUS]
    - [FINDING 1 , STATUS]
    ...
 - role : "Doctor"
   msg : [MSG]
   findings : null
 - role : "Patient"
   msg : [MSG]
   findings :
    - [FINDING 0 , STATUS]
    - [FINDING 1 , STATUS]
    ...
...
--End Format--

Chat:
\end{lstlisting}

\begin{lstlisting}[language=promptlanguage,float=*,caption=Prompt for generating chats with claude. ,label={prompt:chat_claude}]
Create a history taking conversation where a patient has the following medical profile, and is chatting with a doctor.
Include only the history taking part of the chat, do not include any diagnosis or include the doctor confirming the findings.
Be sure to use language appropriate to the patient's profile, and do not make the patient text technical unless indicated.  Use informal terms when appropriate.
Ensure all findings in the findings list are explicitly discussed, both present and absent statuses.

You will generate one turn of messages at a time.  A turn of messages is defined as a Doctor asking a question and a patient response.
In the first turn of messages, there should be the following;
The first message of the first turn should consist of a generic system message which  explicitly states the patient's name, age, gender (stated directly), race if present.  Do not include any findings beyond that list.
The second message of the first turn should consist of a generic greeting from the doctor.  The doctor does not have any information at this point.
The third message of the first turn should consist of the first findings they report.

All previous turns should consist of a provider question and a patient response.

At the end of the chat, all findings must be included in a previous message.

Do not include more than two findings in a question answer pair.
The patient is also unlikely to be able to explain their condition in a coherent manner.  Do not assume that they will clearly offer up the findings, and the doctor may have to ask multiple questions before they get to a finding.
The doctor should not ask leading questions.
Do not assume that the doctor can see the patient. Express everything only through text.

When available, a definition and a previous message from a different patient. for each finding is provided.  Follow the definition of the finding provided.
You must create a different message - including the content - than the previous message if possible given the definition.
---Example---
tongue protrusion with marked deviation:
  definition: upon protrusion of the tongue, there is significant digression away from the midline, toward either the left or right side
  previous message: Yes, actually. I'd say its towards the right side of my tongue.
  status: present
The next message should be discuss one of the other options in the definition (e.g., left side).  For example, "Hmm, maybe on the left side of my tongue."
---End Example---

Disease: {{disease}}

--Patient Profile--
{{demographics_str}}
--End Profile--

Output format in yaml
--Format--
msg_turn:
[CONDENSED]
--End Format--
\end{lstlisting}

\begin{lstlisting}[language=promptlanguage,float=*,caption=Prompt for generating differential diagnosis.  Note that the text between the if statements is only included if candidates are provided. ,label={prompt:ddx}]
The following is a history taking conversation between a patient and a doctor.

Do the following steps;
First,  generate 5 differential diagnoses that are likely given the history taking chat.  For each, briefly discuss the reasoning. Ignore the candidate diagnoses for this step. Output the result in a section beginning with "--First pass--" and ending with "--/First pass--"

{% if candidate_diagnoses -%}

Second, consider the following candidate diagnoses.   You are not required to use them, and should discard them if not appropriate.  For each, briefly discuss the reasoning. Output the result in a section beginning with "--Candidate pass--" and ending with "--/Candidate pass--"

--Candidate Diagnoses--
{{candidate_diagnoses}}
--End Candidate Diagnoses--

Finally, consider the diagnoses from 1) and 2), and pick the 5 most appropriate ones.  You must discard any that aren't appropriate, either from step 1 or 2.

{% else %}
Finally, consider the diagnoses from 1) and pick the 5 most appropriate ones.  You must discard any that aren't appropriate.

{% endif -%}

Output in yaml described at the end, where [DDx #] is replaced with the disease name.
Include a likelihood rating of high, medium, or low in [LIKELIHOOD].  Include a short description of your reasoning in [REASON #].


--Chat--
{{chat_formatted}}
--End Chat--

--Begin Output--
differential_diagnosis:
  [DDx 0]:
      likelihood: [LIKELIHOOD]
      reasoning:
        - "[REASON 0]"
        - "[REASON 1]"
        ...
  [DDx 1]:
      likelihood: [LIKELIHOOD]
      reasoning:
        - "[REASON 0]"
        - "[REASON 1]"
        ...
--End--
\end{lstlisting}

\begin{lstlisting}[language=promptlanguage,float=*,caption=Prompt for judging binary differences between differential diagnoses \cite{tu2024conversationaldiagnosticai}. ,label={prompt:judge_binary}]
Is our predicted diagnosis correct (Y/N)? It is okay if the predicted diagnosis is more specific/detailed.
Predicted diagnosis: {{prediction}}, True diagnosis: {{gold_label}}
Answer [Y/N]:
\end{lstlisting}

\begin{lstlisting}[language=promptlanguage,float=*,caption=Prompt for judging relative differences between differential diagnoses \cite{tu2024conversationaldiagnosticai}. ,label={prompt:judge_classification}]
How close did the differential diagnosis (DDx) come to including the DIAGNOSIS from the answer key?


DDX: {{ddx}}
DIAGNOSIS: {{final_diagnosis}}

CHOICES: 
-  (Unrelated): Nothing in the DDx is related to the diagnosis.
- (Somewhat Related): DDx contains something that is related, but unlikely to be helpful in determining the  diagnosis.
- (Relevant): DDx contains something that is closely related and might have been helpful in determining the  diagnosis.
- (Extremely Relevant): DDx contains something that is very close, but not an exact match to the  diagnosis.
- (Exact Match): DDx includes the  diagnosis.

Choice: Best from the CHOICES
Chosen Dx: diagnosis in DDX which was matched to DIAGNOSIS
location of Dx: location of the chosen Dx in DDX
Rationale: rationale for the choice
\end{lstlisting}

\end{document}